\documentclass[letterpaper, 10 pt, conference]{ieeeconf}  

\IEEEoverridecommandlockouts                              

\usepackage{mathtools}
\usepackage{gensymb}
\usepackage[export]{adjustbox}
\usepackage{graphics} 
\usepackage{epsfig} 
\usepackage{times} 
\usepackage{amsmath} 
\usepackage{tabularx}
\usepackage{amssymb}  
\usepackage{tablefootnote}
\usepackage[table, dvipsnames]{xcolor}
\usepackage{booktabs}
\usepackage{booktabs}
\usepackage{multirow}
\usepackage{multicol}
\usepackage{graphicx}
\usepackage{color, colortbl}
\usepackage[font=small]{caption}
\usepackage{xfrac}
\usepackage{subcaption}     

\usepackage{adjustbox}

\usepackage[backend=biber, backref=true]{biblatex}
\usepackage{hyperref}
\usepackage{amsmath,amssymb,amsfonts}
\usepackage{graphicx}
\usepackage{textcomp}

\DeclareMathOperator*{\argmin}{arg\,min}
\usepackage{authblk}
\let\labelindent\relax
\usepackage{enumitem}
\usepackage{hyperref}
\usepackage{breqn}
\usepackage{algpseudocode}
\usepackage[linesnumbered,ruled,lined]{algorithm2e}

\usepackage{array}
\usepackage{makecell}
\usepackage{tabularray}
\title{\LARGE \bf
 LeGo-Drive: Language-enhanced Goal-oriented \\ Closed-Loop End-to-End Autonomous Driving
 \\
 \vspace{0.30em}
\large{\href{https://reachpranjal.github.io/lego-drive}{\textcolor{JungleGreen}{\texttt{https://reachpranjal.github.io/lego-drive}}}}
}

\author{\textcolor{Blue}{Pranjal Paul$^{1}$, Anant Garg*$^{1}$, Tushar Choudhary*$^{1}$, Arun Kumar Singh$^{2}$, K. Madhava Krishna$^{1}$}
\\ 
$^{1}$\href{https://robotics.iiit.ac.in/}{\textcolor{magenta}{The International Institute of Information Technology, Hyderabad}},
$^{2}$\href{https://tuit.ut.ee/en/content/arun-kumar-singh}{\textcolor{magenta}{University of Tartu, Estonia}}

\thanks{* Equal contribution.}
}

\newcommand{\coolname}{\textit{LeGo-Drive} }

\makeatletter
\let\@oldmaketitle\@maketitle
\renewcommand{\@maketitle}{\@oldmaketitle
\centering
\includegraphics[width=\linewidth]{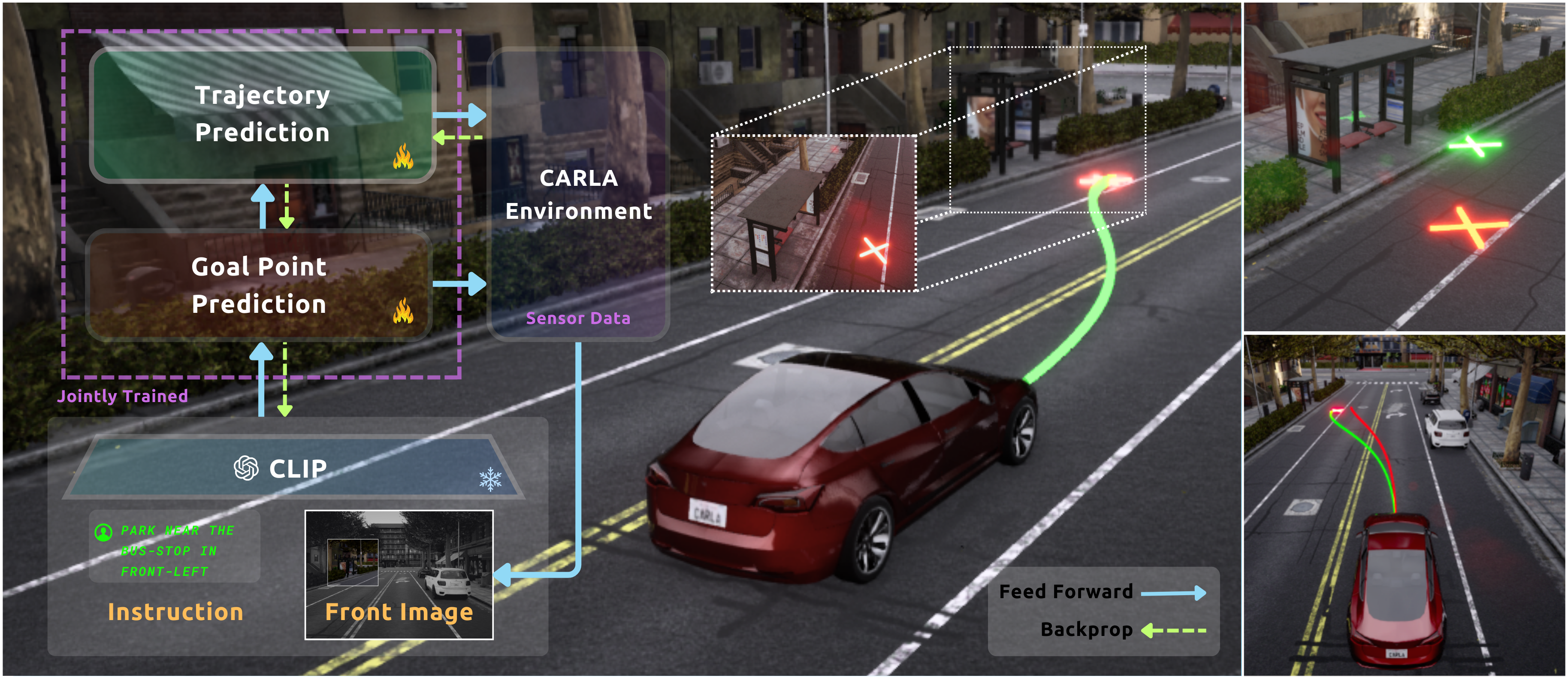}
\captionof{figure}{
The proposed method, \coolname estimates a goal location queried with a navigation instruction- \textit{Park near the bus stop on the front-left} on a single front-facing camera image and coupled it with a differentiable optimizer-based planner that jointly optimizes the trajectory and the goal location. \textit{(Left)} The proposed architecture is shown along with the gradient flow for joint end-to-end training. \textit{(Right-Top)} Goal improvement from initial estimation in \textcolor{Green}{Green} to improved location in \textcolor{red}{Red}. \textit{(Right-Bottom)} Trajectory output in \textcolor{Green}{Green} leading to the improved goal location, compared with the trajectory generated by baseline in \textcolor{red}{Red}.
}
\label{fig:teaser}
\vspace{-0.4cm}
}
\makeatother

\begin{document}

\maketitle
\thispagestyle{empty}
\pagestyle{empty}



\begin{abstract}

Existing Vision-Language models (VLMs) estimate either long-term trajectory waypoints or a set of control actions as a reactive solution for closed-loop planning based on their rich scene comprehension. However, these estimations are coarse and are subjective to their ``world understanding" which may generate sub-optimal decisions due to perception errors. 

In this paper, we introduce \textit{LeGo-Drive}, which aims to address this issue by estimating a goal location based on the given language command as an intermediate representation in an end-to-end setting. The estimated goal might fall in a non-desirable region, like on top of a car for a parking-like command, leading to inadequate planning. Hence, we propose to train the architecture in an end-to-end manner, resulting in iterative refinement of both the goal and the trajectory collectively. We validate the effectiveness of our method through comprehensive experiments conducted in diverse simulated environments. We report significant improvements in standard autonomous driving metrics, with a goal reaching Success Rate of $81\%$. We further showcase the versatility of \coolname across different driving scenarios and linguistic inputs, underscoring its potential for practical deployment in autonomous vehicles and intelligent transportation systems.
\vspace{0.2cm}


\end{abstract}

\section{INTRODUCTION}\label{sec:introduction}

Language-augmented autonomous driving has gained a remarkable surge of interest in recent years. With the advent of Large Language Models (LLMs), existing methods can make informed decisions based on their scene comprehension capability and deliver high-level driving assistance \cite{shao2023lmdrive, tan2019lxmert, sima2023drivelm}. Broadly, there are two classes of approaches. The first approach pertains to reactive behaviour that maps language commands (like \emph{go-fast}, \emph{slow} etc.) to control actions (like throttle, brake and steering commands). The second approach maps higher-level navigation instructions (like \emph{changing lanes}, \emph{overtake} etc.) to long-term trajectory prediction \cite{mao2023gpt, pan2024vlp}. However, in the former, long-term behaviour would require a complex combination of braking, steering, and acceleration primitives. We aim to explore the latter in this study.


Trajectory prediction subjective to the ``logical abilities" of Vision-Language Models (VLMs) may lead to infeasible output due to missed detections or false positives. Instead, a simpler approach would be to gain an intermediate representation that is interpretable and lowers the perception dependency. On this line of thought, we aim to explore a goal-oriented approach where we first map high-level language commands to a desired goal and then subsequently, generate a navigable trajectory towards it. The advantages of such an approach are three-fold. First, the dataset need to be annotated for matching language commands to just goal positions. Moreover, our results are based on providing the supervision of only a very coarse goal region conditioned on the language command, which is easier to obtain compared to the demonstration of the complete driving trajectory. Second, as discussed in \cite{albrecht2021interpretable, ghoul2023interpretable, gu2021densetnt}, goal-directed planning improves the explainability in autonomous driving. Finally, predicting just the goal position would allow the use of smaller and lightweight networks and consequently less data for training, plus faster inference time.

The core challenge in performing language-conditioned goal prediction is that the network should be aware of the vehicle and the scene constraints i.e., predicting a goal that is outside the drivable area is undesirable \footnote{It is possible that the planner handles infeasible goal prediction by stopping short of the predicted position. However, such post-hoc corrections add more burden on the planning and also defeats the explainability objective of our approach.}. For eg., in Figure \ref{fig:teaser}, the command \textit{``Park near the bus stop on the front-left"} provides an initial goal prediction which is at the curb edge which makes the location unreachable. In other words, the language-conditioned goal prediction should be aware of the capabilities of the downstream planner. Our proposed work provides a systematic solution to this end. The core innovations and their benefits are summarized below:

\textbf{Algorithmic Contribution:} We present a lightweight Visual Language Network (VLN) augmented with a parameterized differentiable optimization layer that acts as the downstream planner. Due to the feedback of the planner during training, the VLN network learns to predict goals that are reachable under vehicle kinematic and environment scene constraints. We show that our entire pipeline can be trained by just providing very coarse supervision of goal regions conditioned on the language commands. This is achieved by augmenting the overall loss function with the downstream planning loss that depends on the states of the neighbouring vehicles and the ego's kinematic capabilities. We show that our end-to-end training can not only learn to predict feasible goal positions but also the parameters of the planner that accelerate its convergence to a feasible solution. This contributes to the applicability of our approach with better explainability and faster inference time.

\textbf{Prior Art:} We improve upon the existing literature in two aspects. We demonstrate the impact of our end-to-end training in feasible goal prediction by comparing it against baseline GLC \cite{Rufus_2021}, which does not take the downstream planner into account during training. Second, we also compare our approach against post-hoc corrections, wherein ST-P3 \cite{hu2022stp3} (prior contribution in end-to-end motion planning), tries to reach an infeasible goal prediction generated by the baseline network (Refer Table \ref{tab:model-compare}).

To summarize, our key contributions are:

\begin{enumerate}
    \item A novel planning-guided end-to-end LLM-based goal-point navigation solution that predicts and improves the desired state by dynamically interacting with the environment and generating a collision-free trajectory. 

    \item We conduct extensive closed-loop experiments with different intricate instructions to test the efficacy of the proposed model under different simulation environments with different lighting and weather conditions.
\end{enumerate}



\section{RELATED WORK} \label{relatedwork}

\subsection{Visual Grounding}

Visual grounding aims to associate a natural language query with the most relevant visual elements or objects in a visual scene. Visual grounding tasks were previously approached as referring expression comprehension (REC), 
which generates region proposals and then exploits the language expression to select the best-matching region.
Conversely, one-stage methods also known as Referring Image Segmentation (RIS)\cite{liao2020realtime, yang2020improving}, integrate linguistic and visual features within a network and directly predict the target box \cite{deng2022transvg, liao2020realtime}.
\cite{Rufus_2021} uses the RIS approach for the task of identifying navigable regions on the drivable areas based on a language command. However, the work is limited to scene understanding and does not include navigation simulations, as trajectory planning relies on precise goal-point location, which they do not address.

\subsection{End-to-End Autonomous Driving}
In recent years, End-to-End learning-based research has been a prominent focus. The E2E approach is a unified data-driven learning-based approach, to ensure safe motion planning in contrast to conventional rule-based designs that optimize each task in a disjoint fashion instead of optimizing for a unified target leading to compounding errors. UniAD \cite{hu2023_uniad} is a current state-of-the-art method on nuScenes dataset \cite{caesar2020nuscenes} which uses rasterized scene representation to identify vital components within the P3 \cite{sadat2020perceive} framework. ST-P3 \cite{hu2022stp3} was a prior art that explored the interpretability of the vision-based end-to-end ADS. Due to computational constraints, we choose ST-P3 as our baseline for motion planning over UniAD.

\subsection{Planning-oriented Vision-Language Navigation}

LLMs have shown promising results in the Autonomous Driving System (ADS) domain for their multimodal understanding and natural interaction with humans. Existing works \cite{mao2023gpt, sha2023languagempc, xu2023drivegpt4, pan2024vlp} use LLM to reason driving scenes and predict control inputs. However, they are limited to open-loop settings. More recent works \cite{chen2023driving, shao2023lmdrive, sima2023drivelm} focus on adapting to closed-loop solutions. They either directly estimate the control actions or map them to a set of discrete action spaces. These are coarse and are susceptible to perception errors due to their heavy reliance on VLMs knowledge retrieval capabilities that may generate non-smooth motion for intricate cases like parking, highway merge, etc. which require complex combinations of control actions.

\section{DATASET} \label{dataset}

In this section, we elaborate upon our dataset creation and annotation strategy tailored to our requirement to develop an intelligent driving agent that incorporates vision-centric data sourced from the CARLA simulator and coupled with navigation instructions. We assume that the agent is employed with the necessary privileged information to execute a successful closed-loop navigation. 
\vspace{0.3cm}

\textbf{Dataset Overview:} Prior works, such as the Talk2Car dataset \cite{Deruyttere_2019}, primarily focus on scene understanding by annotating bounding boxes for object references. Further works, such as Talk2Car-RegSeg \cite{Rufus_2021} aim to include navigation by annotating segmentation masks for navigable regions. We expand upon these datasets by encompassing a wide variety of driving maneuvers, including lane changes, speed adjustments, turns, passing or stopping for other objects or vehicles, navigating through intersections, and stopping at crosswalks or traffic signals, on which we later demonstrate closed-loop navigation. The created \coolname dataset comprises 4500 training and 1000 validation data points. We present results, baseline comparisons, and ablations using both complex and simpler command annotations.
\vspace{0.3cm}

\textbf{Simulation Setup:} The \coolname dataset collection procedure consists of two stages: 1) synchronous recording of the driving agent state with camera sensor data, followed by traffic agents, and 2) parsing and annotating the collected data with navigation directives. We record data at 10 FPS and to avoid redundancy between consecutive frames, data points is filtered at a distance interval of 10m. For each frame, we collect the ego-agent's states, i.e. position, and velocity, ego-lane with a 50-meter range in both front \& rear directions, front RGB camera image, and traffic agent states (position and velocities) utilizing the rule-based expert agent, all in ego-frame. 
The dataset is diverse across 6 different towns covering a variety of distinct environments representing various driving scenarios with different lane configurations, traffic densities, lighting, and weather conditions. Additionally, the dataset includes a variety of objects commonly observed in outdoor scenes such as bus stops, food stalls, and traffic signals.

\begin{table}[!h]
\centering
\begin{tabular}{l|p{0.6\linewidth}}
\toprule
\textbf{Command type} & \textbf{Example Prompts} \\
\midrule
\multirow{3}{*}{\textbf{I. Object-centric}} & Park behind the bike in front \\
                                & Slow down at the food stall on the right \\
                                & Turn left from the gas station \\
\midrule
\multirow{3}{*}{\textbf{II. Lane maneuver}} & Maintain the same lane \\
                               & Take right at the next intersection \\
                               & Switch to the left lane \\
\midrule
\multirow{3}{*}{\textbf{III. Composite}} & Turn right and head to the cafe \\
                           & Keep a safe distance to the car in front \\
                           & Switch lane and stop at the parking area \\
\bottomrule
\end{tabular}

\caption{Samples of navigation instruction from the \coolname dataset}
\label{tab:prompts}
\vspace{-1em}
\end{table}

\textbf{Language Command Annotation:} Each frame is labeled manually with proper navigation commands corresponding to goal region segmentation masks to cover a range of driving scenarios. We consider 3 different command categories: 1). \textit{Object-centric commands}, which directly refer to an object visible in the current camera frame, 2). \textit{Lane Maneuvering commands}, which are instructions that are specific to actions related to a lane change or adjusting within the lane, and 3). \textit{Composite commands}, which connect multiple instructions to simulate real driving scenarios. We utilize ChatGPT API to generate different variants with similar semantic meanings. Table \ref{tab:prompts} shows a few samples of example instructions from our dataset. It bears noting that we do not incorporate handling misleading instructions. This capability is imperative in scene reasoning models which may be considered for future expansion; however, it falls outside the scope of our current study.

\section{METHODOLOGY} \label{methodology}

\coolname is a framework devised to address the feasibility of coarse estimation of control actions from VLAs, treating it as a short-term goal-reaching problem. This is achieved through the learning of trajectory optimizer parameters together with behavioural inputs by generating and improving a feasible goal aligned with the navigation instruction.

\begin{figure*}[!t]
\centering
\includegraphics[width=\textwidth]{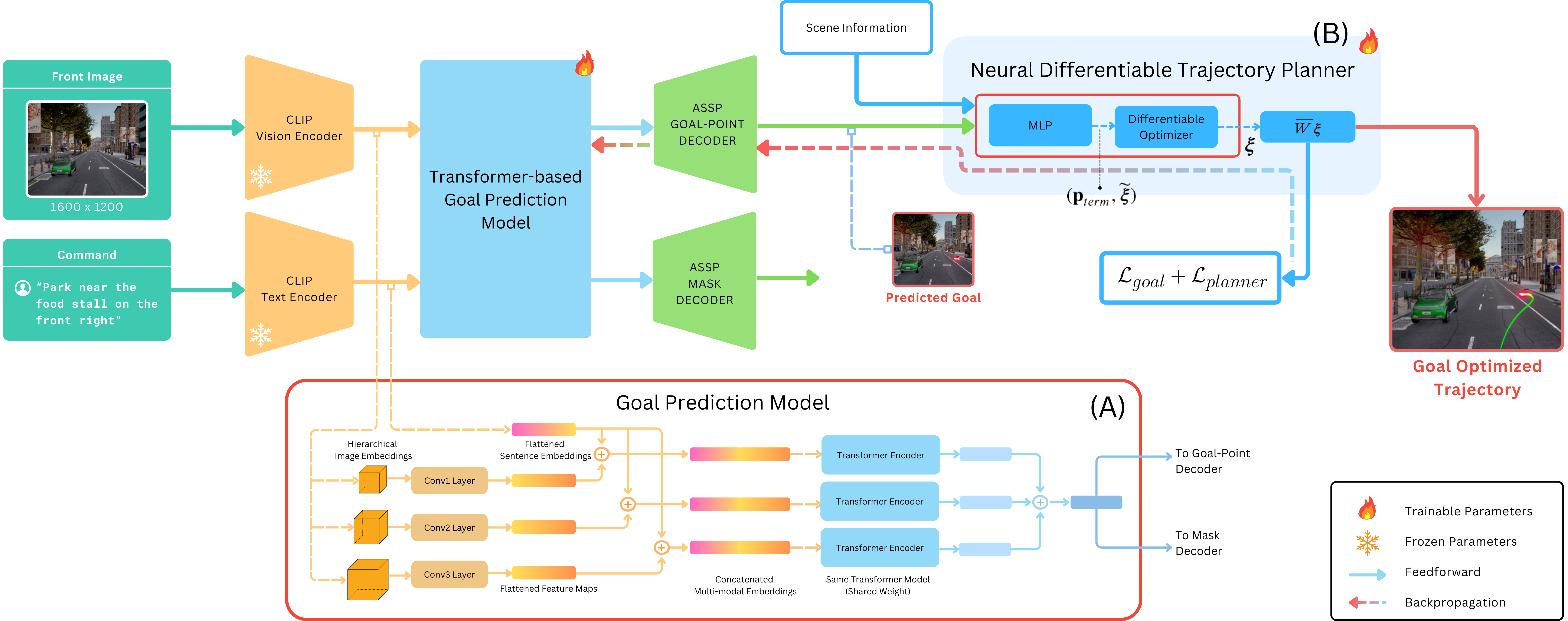}
\caption{\textbf{\coolname Architecture:} Our architecture comprises of two modules: \textit{(A) Goal Prediction} module and \textit{(B) Differentiable Trajectory Planner}. We propose the advantage of end-to-end training for combined goal and trajectory improvement, for which the gradient-flow is clearly shown. (Refer to Section IV-B for trajectory variable definition)}
\label{fig:architecture}
\end{figure*}

As illustrated in Figure \ref{fig:architecture}, the architecture is composed of two major segments:
\begin{enumerate}
    \item \textit{Goal Prediction} module that accepts a front-view image $\mathbf{I} \in \mathbb{R}^{H \times W \times 3}$ and a corresponding language command $\mathbf{L} = { \{ l_0, \dots, l_k \} }$, where $l_i$ is a word token and $k$ is the length of the command; to generate or predict a segmentation mask $\mathbf{M} \in \mathbb{R}^{H \times W}$ followed by a goal location $\hat{\mathbf{g}}_i \in \mathbf{M}$, and, 
    
    \item \textit{Differentiable Planner} that generates a trajectory $\mathbf{T} \in {\mathbb{R}^{N \times 2}}$ which is jointly optimized for the estimated goal and trajectory optimizer parameters resulting in improvement of the desired position coordinates $\hat{\mathbf{g}}_i$ to a navigable location $\hat{\mathbf{g}}_i^\ast \in {\mathbb{R}^2}$ when trained end-to-end.
\end{enumerate}

\subsection{Goal Prediction Module}
To encode the given navigation command, we tokenize the linguistic command using CLIP \cite{radford2021learning} tokenizer and pass it through
CLIP text encoder to obtain text embeddings \textbf{T}. To get the image features from the given front camera image, we utilize the CLIP image encoder with ResNet-101 backbone. Hierarchical features are known to be beneficial for semantic segmentation; hence, we extract different visual feature $\mathbf{V_i} \in \mathbb{R}^{C_{i} \times H_{i} \times W_{i}}$ where \(i \in \{2, 3, 4\}\) after the $2^{nd}$, $3^{rd}$, and $4^{th}$ layers of the ResNet backbone. Each $\mathbf{V_i}$ is passed through convolutional blocks \(\text{ConvBlock}_i\) to bring them into a standard size with equal channel sizes, heights, and widths.

To capture the multi-modal context from the image and text features, we further use a transformer encoder adopted from the DETR \cite{carion2020endtoend} architecture. All features $\textbf{T}$, $\mathbf{V_2}$, $\mathbf{V_3}$, $\mathbf{V_4}$ are flattened, and text features are concatenated with different $\mathbf{V_i}$ individually to get multimodal features $\mathbf{M_i} = \mathbf{V_i} \oplus \mathbf{T}$. The $\mathbf{M_i}$ is then individually passed to the transformer encoder where the multi-headed self-attention layer helps in cross-modality interaction between the different kinds of features to obtain $\mathbf{X_i}$ as the encoder output with the same shape as $\mathbf{M_i}$.

We have two decoder heads, one each for the segmentation mask prediction and the goal point prediction task respectively. To predict the segmentation mask, $\mathbf{X_i}$ undergoes further reshape and restructure operations to reshape it into $\mathbb{R}^{C \times H \times W}$, resulting in $\mathbf{Z_i}$. For the segmentation mask prediction, we stack the $\mathbf{Z_i}$ from all layers to shape $\mathbb{R}^{C+C+C \times H \times W}$. 

Both prediction heads use ASPP decoders from \cite{chen2018encoderdecoder}. For segmentation mask prediction, ASPP outputs pass through a convolutional upsampling block that includes bilinear upsampling at specified stages to increase spatial resolution. The output finally undergoes sigmoid activation to produce binary masks. In the goal point prediction decoder, it consists of convolutional layers followed by fully connected layers with the output reshaped to $\mathbb{R}^{2 \times 1}$ representing a pixel location on the image.

First, the segmentation mask prediction head is trained end-to-end with BCE loss between the predicted segmentation mask and the human-annotated ground truth segmentation mask. After a few epochs, the goal point prediction head is trained similarly end-to-end with a smooth L1 loss between the predicted goal point and human annotated ground truth goal point.

\textit{Complex Commands and Scene Understanding}: To handle composite instructions serving cases where the final goal location is not visible in the current frame, we adapt our approach by decomposing the complex command into a list of atomic commands that need to be followed sequentially. For example, "switch to the left lane and then follow the black car" can be decomposed into "switch to the left lane" and "follow the black car". To decompose such complex commands, we construct a list of atomic commands $L$, covering a wide range of simple actions such as lane changes, turns, speed adjustments, and object references. Upon receiving a complex command, we utilize the few-shot learning technique to prompt an LLM to decompose the given complex command into a sequential list of atomic commands $l_i$, from $L$. These atomic commands are then executed iteratively with our pipeline, with the predicted goal-point location serving as intermediate waypoints to help us reach the final goal point.

\subsection{Neural Differentiable Planner}
Our planner takes the shape of an optimization problem that is embedded with learnable parameters to improve the downstream task of following the goal generated by the VLN and accelerate the convergence. In the following, we first introduce the basic structure of our trajectory optimizer followed by its integration with a network.

\textbf{Basic Problem Formulation:} We assume access to the lane centre-line and use it to construct the Frenet Frame \cite{werling2010optimal}. The trajectory planning is formulated in this frame and has the advantage that the longitudinal and lateral motions of the car are aligned with the $X$ and $Y$ axis of the Frenet frame respectively. With this notation in place, our trajectory optimization problem has the following form:

\small
\begin{subequations}
\begin{gather}
  \min  \sum_k \dddot{x}[k]^2+\dddot{y}[k]^2\label{cost} \\
    (x^{(r)}[0],  y^{(r)}[0]) = \textbf{b}_0, (x^{(r)}[n],  y^{(r)}[n]) = \textbf{b}_f \label{boundary_cond}\\
    g_i(x^{(r)}[k], y^{(r)}[k])\leq 0 \label{ineq_constraint}
\end{gather}
\end{subequations}
\normalsize
\vspace{-0.5cm}
\small
\begin{subequations}
\end{subequations}
\normalsize

\noindent where $(x[k], y[k]$ represents the position of the ego-vehicle at time step $k$. The cost function penalizes high magnitudes of jerk. The equality constraints \eqref{boundary_cond} ensure that the planned trajectory satisfies the initial and final boundary conditions on the $r^{th}$ derivative of the planned trajectory. We use $r= \{0, 1, 2\}$ in our formulation. The inequality constraints \eqref{ineq_constraint} also depend on the derivatives up to the $r^{th}$ order and include velocity, acceleration, and lane bounds along with constraints on collision avoidance and curvature. The algebraic structures of $g_i(.)$ are taken from our prior work \cite{shrestha2023end}.

To ensure that we optimize in the space of smooth trajectories, we parameterize the motions along the $X-Y$ directions in the following form:

\small
\begin{gather}
    \begin{bmatrix}
        x[0],x[1], \dots, x[k] 
    \end{bmatrix} = \textbf{W}\textbf{c}_{x},
    \begin{bmatrix}
        y[0], y[1], \dots, y[k] 
    \end{bmatrix} = \textbf{W}\textbf{c}_{y}, \\
    \overline{\textbf{W}} = 
    \begin{bmatrix}
        \textbf{W} & \textbf{0} \\
        \textbf{0} & \textbf{W}
    \end{bmatrix}
    \label{param}
\end{gather}
\normalsize

\noindent Using \eqref{param}, the optimization \eqref{cost}-\eqref{ineq_constraint}, can be written in the following compact form

\begin{subequations}
\begin{gather}
    \boldsymbol{\xi}^* = \argmin_{\boldsymbol{\xi}} \frac{1}{2}\boldsymbol{\xi}^T\textbf{Q}\boldsymbol{\xi}+\textbf{q}^T(\textbf{p})\boldsymbol{\xi}, \label{lower_cost_reform} \\
    \textbf{A}\boldsymbol{\xi} = \textbf{b} (\textbf{p}_{term})\label{lower_eq_reform} \\
    \textbf{g}(\boldsymbol{\xi}) \leq 0\label{lower_ineq} \\
    \widetilde{\textbf{A}} \boldsymbol{\xi} = \widetilde{\boldsymbol{\xi}} \label{partial_sol}
\end{gather}
\end{subequations}

\noindent where $\boldsymbol{\xi} = (\textbf{c}_x, \textbf{c}_y)$ and $\textbf{p}_{term}$ is the collection of terminal positions and velocities along the $x$ and $y$ component of motion. The matrix $\widetilde{\textbf{A}}$ and the constant vector $\widetilde{\boldsymbol{\xi}}$ is not part of the original trajectory optimization problem but has been introduced due to some specific reason discussed below.

\textbf{Conditioning on the Partial Solution}: Optimization \eqref{lower_cost_reform}-\eqref{lower_ineq} can be challenging to solve in real-time due the presence of non-convex constraints $\textbf{g}$. Inspired by \cite{donti2021dc3}, we explore a possible solution for accelerating the convergence of the optimizer. Imagine that, we are given a partial solution to the problem. That is, we have some of the components of the solution vector $\boldsymbol{\xi}$ which we denote as $\widetilde{\boldsymbol{\xi}}$ in \eqref{partial_sol}. We want to use such information to bias the solution process towards favourable regions. To this end, we introduce explicit conditioning through the affine constraints \eqref{partial_sol}, wherein matrix $\widetilde{\textbf{A}}$ is simply some specific rows of an identity matrix. More precisely, imagine that $\widetilde{\boldsymbol{\xi}}$ is formed by first ten elements of $\boldsymbol{\xi}$, then $\widetilde{\textbf{A}}$ will be formed by extracting the first ten rows of the identity matrix of size equal to dimension of ${\boldsymbol{\xi}}$.

\textbf{Neural Planners:} We want to learn the terminal states $\mathbf{p}_{term}$ and $\widetilde{\boldsymbol{\xi}}$ in end-to-end fashion along with our VLN network. To this end, we present a hybrid planning network that consists of a multi-layer perceptron (MLP) and differentiable optimization layer embedded with \eqref{lower_cost_reform}-\eqref{partial_sol}. (Refer Figure \ref{fig:architecture}-B)

\subsection{Training and End-to-End Framework}
The novelty of our method stems from its modular end-to-end planning framework wherein the framework optimizes the goal prediction module and prioritizes trajectory optimization while ensuring that the acquired behavioural inputs effectively facilitate optimizer convergence. Fundamental to the architecture is the iterative refinement of differentiable modules, whereby the enhancement of goal prediction positively influences trajectory optimization, and conversely, refined trajectory planning contributes to improved goal prediction. This cyclic progression forms the backbone of our design, ensuring a cohesive and iterative improvement loop within the system.

Due to the modular nature of the architecture, the model can be trained in two ways: 
\begin{enumerate}
    \item \textbf{\textit{LeGo-Drive E2E}:} denotes the joint training of both modules. The model is trained over the combined loss $ \mathcal{L} = \mathcal{L}_{goal} + \mathcal{L}_{planner}$ where goal loss $\mathcal{L}_{goal}$ is the MSE loss calculated between the predicted goal $(x_g, y_g))$ and the endpoint of the predicted trajectory $(\mathbf{x_{-1}}, \mathbf{y_{-1}})$ and planner loss $\mathcal{L}_{planner}$, which is a combination of violation of non-convex constraints $\mathbf{g}$ pertaining to lane offset, collision avoidance and kinematic constraint. The gradient flows from the planner to the goal prediction part as shown in Figure \ref{fig:architecture}.
    \begin{subequations}
    \begin{gather}
        \mathcal{L}_{goal} = \frac{1}{N} \sum_{i=1}^{N} \| \left( \mathbf{x}_{-1}, \;\mathbf{y}_{-1} \right) - \left( {x_g, \;y_g} \right) \|_2^2 \label{goal-loss} \\
        \mathcal{L}_{planner} = || \mathbf{max}\left(0, \; \mathbf{g} \right) ||_2
    \label{planner-loss}
    \end{gather}
    \end{subequations}

    \item \textbf{\textit{LeGo-Drive Decoupled}:} denotes the training process where both the goal prediction module and planner module are trained separately. First, the goal prediction module is trained over MSE loss between the ground truth mask centroid and the predicted goal. Subsequently, the planner is trained on $\mathcal{L}_{planner}$ while keeping the parameters of the goal prediction module frozen.

\end{enumerate}
The end-to-end training  requires backpropagating through the optimization layer modeling the trajectory planning process, which can be done in two ways namely: implicit differentiation and algorithmic unrolling \cite{pineda2022theseus}. The support for the former in terms of existing libraries is mostly restricted to convex problems \cite{agrawal2019differentiable}, or unconstrained non-linear least squares \cite{pineda2022theseus}. In our approach, we build a custom backprogation routine following algorithm unrolling, following our prior work \cite{shrestha2023end}. An advantage of our approach is that it can handle constraints, and the backpropagation can be made free of matrix factorization \cite{shrestha2023end}.

The performance of both methods is shown in Table \ref{tab:model-compare} which is further analyzed in Section \ref{experiments}. 


\section{EXPERIMENTS AND RESULTS} \label{experiments}





\begin{table*}[htp]
\centering
\renewcommand{\arraystretch}{1.5}
\begin{tabular}{lccccccc}\toprule

\multicolumn{2}{c}{\textbf{Model}} & \multicolumn{1}{c}{\textbf{minFDE (m)} $\downarrow$} & \multicolumn{1}{c}{\textbf{Smoothness (avg.)} $\downarrow$} & \multicolumn{1}{c}{\textbf{Success Rate (\%)} $\uparrow$} \\ \midrule
\multicolumn{2}{c}{\textbf{ST-P3 + GLC}} & 0.5145  & 0.1836 & 47.1 \\
\multicolumn{2}{c}{\textbf{\coolname \textit{Dec.} \textit{(Ours)}}} & 0.3982  & 0.1662 & 70.1 \\
\multicolumn{2}{c}{\textbf{\coolname \textit{E2E} \textit{(Ours)}}} & \textbf{0.2985}  & \textbf{0.0603} & \textbf{81.2} \\
\bottomrule
\end{tabular}
\caption{\textbf{Model Comparison:} We compare our proposed approach \textbf{LeGo-Drive-E2E} against the modular decoupled architecture along with architecture formed by combining baselines. Our method excels in all three trajectory metrics evaluated over a validation set of 1000 frames.}
\label{tab:model-compare}
\end{table*}

\subsection{Implementation Details}
\textbf{Perception:} The input to the model is an RGB image of size $1600 \times 1200$ pixels and a language instruction with a maximum sentence length of 20 word tokens. We use different variants of CLIP to extract vision embeddings of various sizes and text embeddings of feature length $1024$, respectively. The model predicts a goal location and a segmentation mask in pixel space with a mask threshold set to 40\%. Additionally, this pixel coordinate is transformed into the egocentric frame for the planner.

\textbf{Planning:} The differentiable optimizer-based planner operates in the road-aligned Frenet frame. We utilize the ego lane as the reference path to transform scene inputs accordingly, adhering to lane constraints based on the simulator settings. Additionally, the vehicle control constraints follow real-world parameters. We use the default controller for closed-loop navigation. The planning horizon is of 6 seconds with a step length of 0.5 meters and considers the 5 nearest obstacles within the range of 50 meters with varying velocities.

\textbf{Training:} The model is trained using Adam optimizer with weight decay of $5e^{-4}$ in a batch size of 16 for 100 epochs with the initial learning rate set to $6e^{-5}$ and polynomial learning rate decay of $0.2$. We have trained and evaluated the model in a single Nvidia RTX 4090 GPU which takes approximately 6 hours to train end-to-end.

\subsection{Evaluation Metrics}
We evaluate \coolname based on the command type with L2 as the primary metric. To consider improvement in goal location, we address: 1). the closeness of the predicted goal w.r.t. to the mask centroid and 2). the lane centre. We also report the proximity to the nearest obstacles agnostic to the command type.

To assess improvement in the trajectory, we evaluate the minFDE (minimum Final Displacement Error) metric adopted from \cite{hu2023_uniad}, defined as the L2 distance between the goal location and the trajectory endpoint. Further, we analyze goal reachability in terms of Success Rate if the ego-vehicle reaches the goal within the radius of 3 meters. Also, we gauge Smoothness, based on how gradually the trajectory approaches the goal, with a slower convergence rate indicating smoother behaviour.

\subsection{Experimental Results}
\subsubsection{Goal Improvement}
Table \ref{tab:goalimproveval} compares the goal evaluation metrics between \textit{LeGo-Drive Decoupled} (Initial) and the proposed \textit{LeGo-Drive E2E} (Improved) for different command types. The E2E approach consistently excels on all the metrics. The model closely approximates the mask centroid which is an ideal location in most scenes evident from the qualitative result. Moreover, it performs comparative w.r.t. the obstacle proximity averaging over variety of driving cases, including parking and obstacle avoidance during lane change. However, with $25\%-30\%$ improvement of the goal location to the lane centre for a single maneuver command, interestingly, the model shows $75\%$ improvement in compound commands which proves the effectiveness of the proposed method. This can be justified by the controlled actions due to the intermediate improved goal corresponding to the first atomic command which compounded to the enhanced performance.


\begin{table}[!h]
\centering
\renewcommand{\arraystretch}{1.5}
\begin{tabular}{lccccccc}
\toprule
\multicolumn{2}{c}{\multirow{2}{*}{\textbf{Type}}} & \multicolumn{2}{c}{\textbf{Obstacle $\uparrow$}} & \multicolumn{2}{c}{\textbf{Mask Centroid $\downarrow$}} & \multicolumn{2}{c}{\textbf{Lane Center $\downarrow$}}  \\ \cmidrule(lr){3-4} \cmidrule(lr){5-6} \cmidrule(lr){7-8}
& & Initial & Improv. & Initial & Improv. & Initial & Improv. \\ 
\midrule
\multicolumn{2}{c}{\textbf{I}} & 5.5141
  &\textbf{6.5067}  & 2.7641   &\textbf{1.2633}  & 2.1443  &\textbf{1.8104} \\
\multicolumn{2}{c}{\textbf{II}}  & 4.7855  &\textbf{5.4258}  & 3.9494   &\textbf{2.0360}  & 3.2525  &\textbf{2.9559} \\
\multicolumn{2}{c}{\textbf{III}}  & 4.5602  &\textbf{5.9259}  & 3.8007   &\textbf{1.4245}  & 1.5358  &\textbf{1.1823} \\
\bottomrule
\end{tabular}
\caption{\textbf{Goal Improvement:} Avg. Distance (in meters) of goal w.r.t. different command types} 
\label{tab:goalimproveval}
\end{table}

\begin{figure}[!h]
    \centering
    \includegraphics[width=0.95\linewidth]{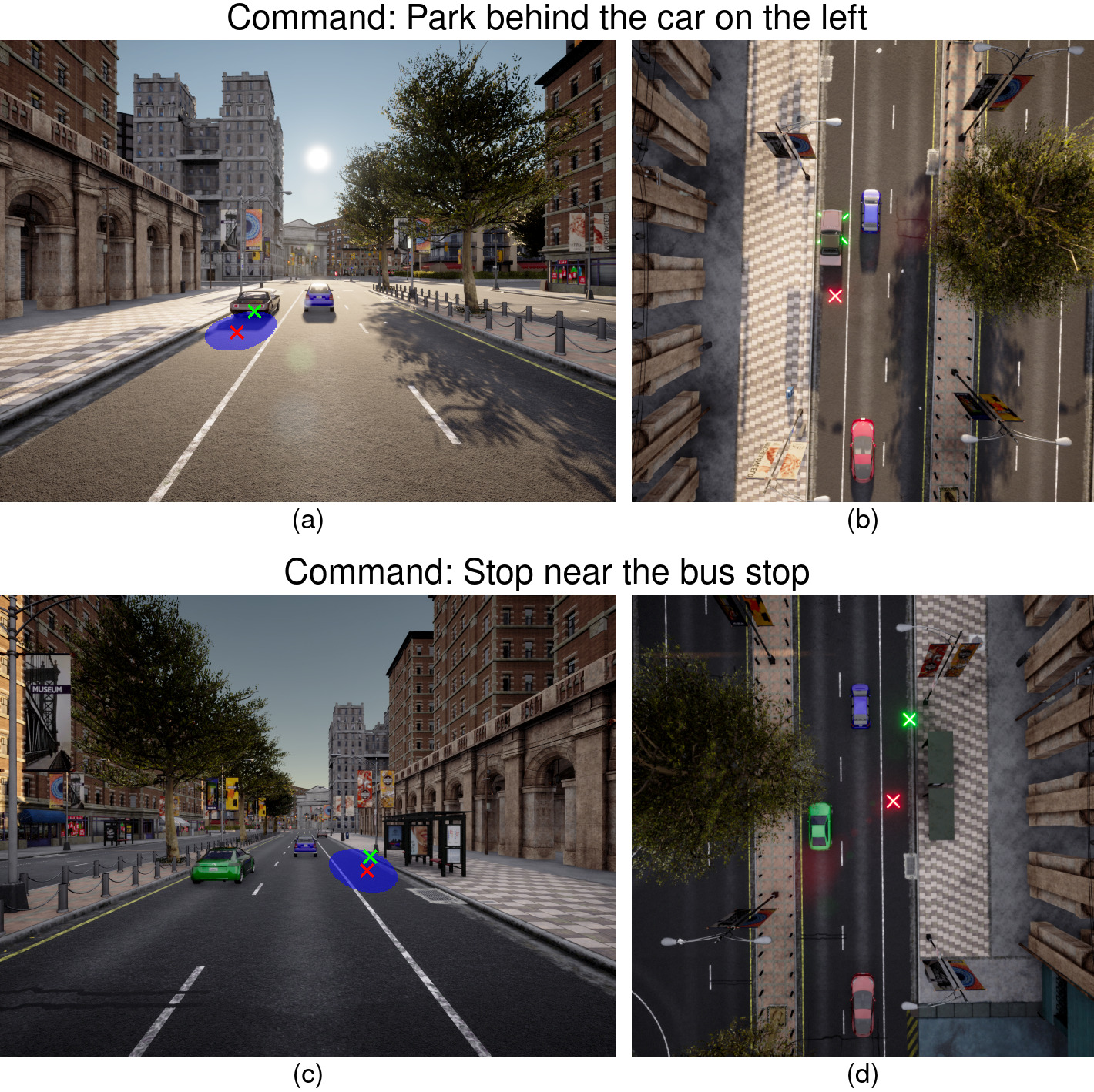}
    \caption{Goal Improvement for different object-centric parking commands. (Left) Front-view image on which command is queried. (Right) Top-down view of the scene. The goal location improves from an undesirable location in \textcolor{Green}{Green} (On top of the car in (a) and at the curb edge in (b)) to a reachable location in \textcolor{red}{Red}}.
    \label{fig:goal-improv-park}
\end{figure}

\begin{figure}[!h]
    \centering
    \includegraphics[width=0.95\linewidth]{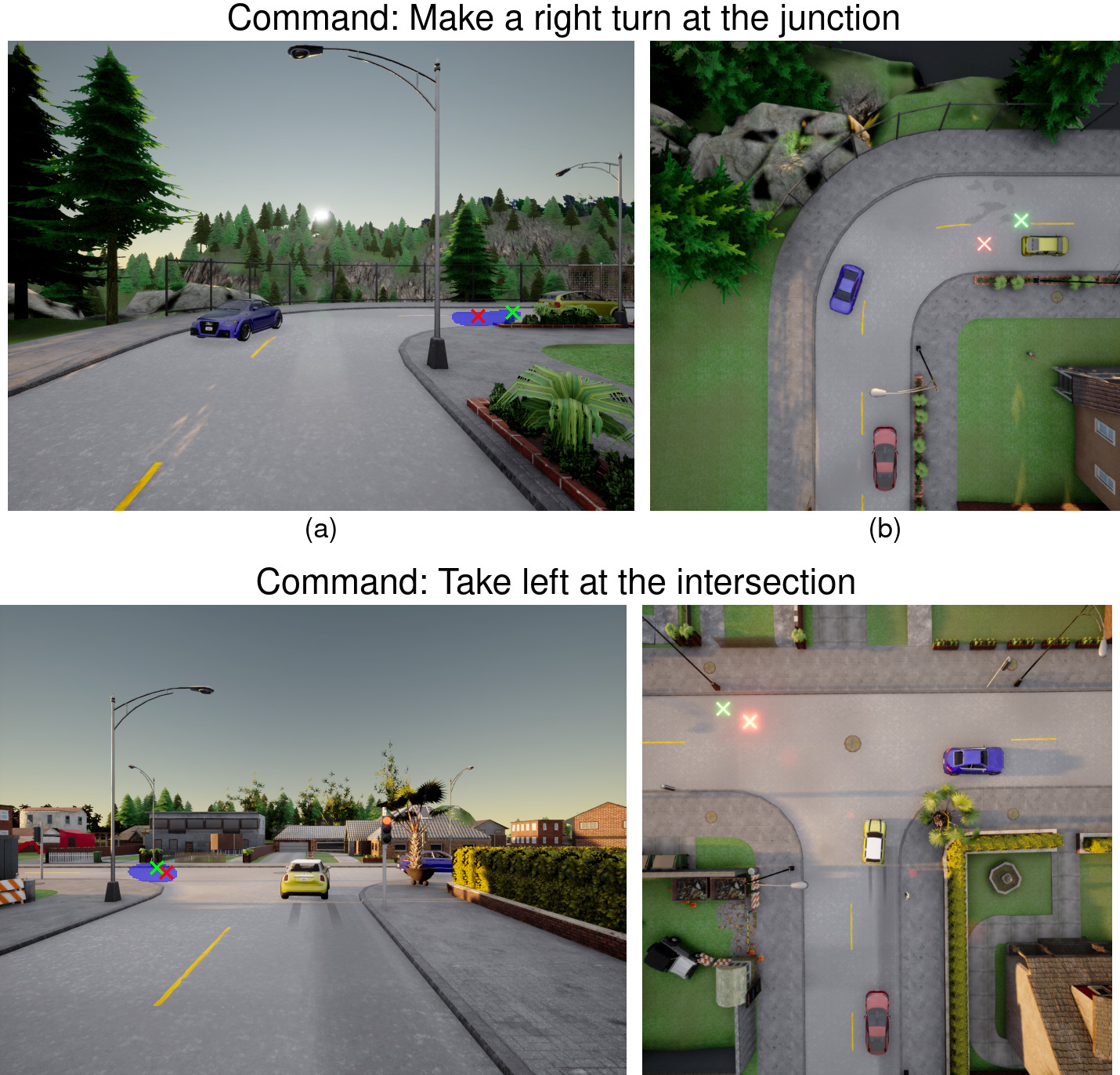}
    \caption{Results for the case of turning commands. In both images (top, bottom), the initial goal in \textcolor{Green}{Green} is at a higher offset from the lane centre. The model approximates the improved version shown in \textcolor{red}{Red} to the lane centre}.
    \label{fig:goal-improv-turn}
    \vspace{-1.5em}
\end{figure}

Figure \ref{fig:goal-improv-park} and \ref{fig:goal-improv-turn} show the qualitative results across various categories of language commands. The visualizations reveal instances where the goal prediction by \textit{LeGo-Drive Decoupled} tends to fall into the infeasible areas. However, through our approach, these goals are subsequently refined and improved.

\subsubsection{Trajectory Improvement Evaluation}


We benchmark the planner performance against ST-P3, a prior art in an end-to-end motion planner domain. For even comparison, we use the improved goal(s) predicted by the trained \textit{LeGo-Drive E2E} model as the desired location. Table \ref{tab:trajeval} reports the results for different commands. The proposed approach, benefiting from both goal and trajectory improvement, surpasses the baseline by a large extent. It clearly ensures goal reachability with a high success rate and smooth collision-free trajectories, spanning across commands based on different driving scenarios. There is a significant decrease in minFDE by $60\%$ for composite command which stems from the basic ideology of the proposed model.

\begin{table}[!h]
\centering
\renewcommand{\arraystretch}{1.5}
\begin{tabular}{cccccccc}\toprule
\multicolumn{2}{c}{\multirow{2}{*}{\textbf{Type}}} & \multicolumn{2}{c}{\textbf{minFDE (m) $\downarrow$}} & \multicolumn{2}{c}{\textbf{Smoothness $\downarrow$}} & \multicolumn{2}{c}{\textbf{SR (\%) $\uparrow$}} \\ \cmidrule(lr){3-4} \cmidrule(lr){5-6} \cmidrule(lr){7-8}
& & ST-P3 & \textit{Ours} & ST-P3 & \textit{Ours} & ST-P3 & \textit{Ours} \\\midrule
\multicolumn{2}{c}{\textbf{I}} & 0.5145 & \textbf{0.4639} & 0.1836 & \textbf{0.0384} & 50.2 & \textbf{79.1} \\
\multicolumn{2}{c}{\textbf{II}} & 0.7710 & \textbf{0.2985} & 0.1679 & \textbf{0.1365} & 36.2 & \textbf{82.4} \\
\multicolumn{2}{c}{\textbf{III}} & 0.6477 & \textbf{0.3982} & 0.1836 & \textbf{0.0603} & 45.4 & \textbf{80.1} \\
\bottomrule
\end{tabular}
\caption{\textbf{Trajectory Evaluation:} Evaluating trajectories based on goal reaching}
\label{tab:trajeval}
\end{table}

\begin{figure}[t]
    \centering
    \includegraphics[width=0.95\linewidth]{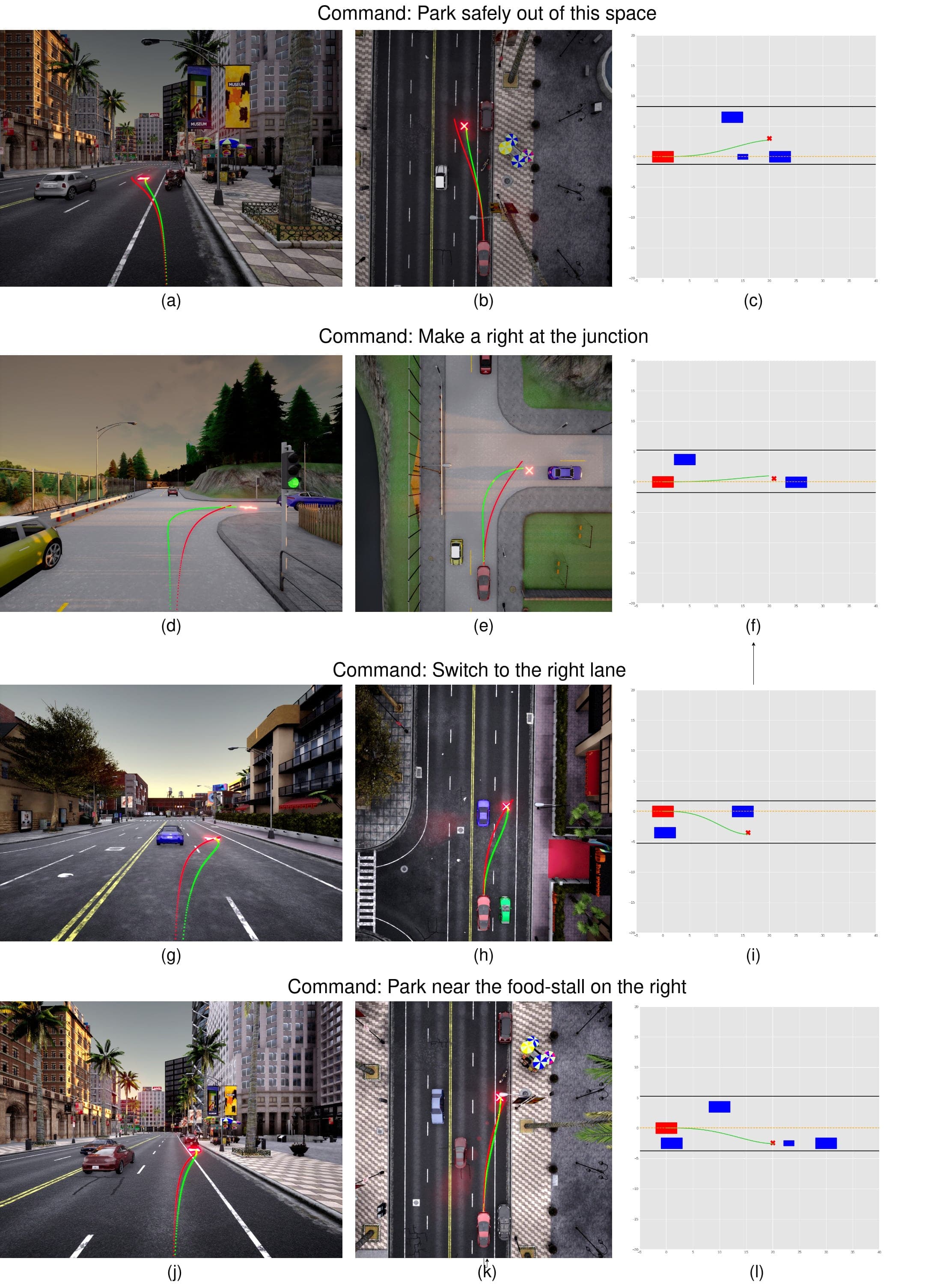}
    \caption{Qualitative Result of the Trajectory Improvement for different navigation instruction leading to an \emph{improved} goal. The baseline ST-P3 trajectory shown in \textcolor{Red}{Red} consistently plans a non-smooth trajectory compared to Ours, shown in \textcolor{Green}{Green}. The third image in all the rows shows our planning in Frenet frame with \textcolor{red}{Red} rectangle as ego-vehicle, \textcolor{blue}{Blue} as surrounding vehicles and \textcolor{red}{Red} cross shows the goal location along with lane bounds in solid Black lines}
    \label{fig:traj-improv}
\end{figure}

Figure \ref{fig:traj-improv} provides a qualitative comparison between trajectories generated by our proposed end-to-end approach and the ST-P3 model across various command categories. The visual comparison highlights that, in contrast to the proposed approach, trajectories from ST-P3 frequently deviate from the intended goal location, often leading to misalignment or incorrect orientations.

\subsubsection{Model Comparison}

We further perform experiments on the following models to address the preferred choice of architecture for our original objective with different commands across different scenes and simulation setups, evaluating on the trajectory improvement metrics:
\begin{itemize}
    \item \textbf{ST-P3 + GLC:} which is developed by cascading two independent architectures, ST-P3 as end-to-end motion planner and GLC as perception module, tailored for their specific tasks. 

    \item \textbf{\textit{LeGo-Drive Decoupled}:} our modular architecture to follow the traditional paradigm similar to the ST-P3 + GLC

    \item \textbf{\textit{LeGo-Drive E2E}:} our proposed approach
\end{itemize}
Based on \ref{tab:model-compare}, our proposed architecture outperforms the baseline by a percentage difference of $~35\%$ in the goal reachability (SR) maintaining smoother convergence. The decoupled version of our model performs relatively well for trajectory improvement with a reasonable difference in Success Rate. However, as previously analyzed in \ref{tab:goalimproveval}, the E2E approach shows superior performance in the goal improvement metrics.

\section{CONCLUSION} \label{conclusion}

Our study reveals a distinct advantage of the proposed end-to-end approach compared to traditional decoupled methods by solving it as a goal-point navigation problem. The joint training of the goal-prediction module with a differentiable optimizer-based trajectory planner highlights the efficacy of our method leading to enhanced accuracy and context-aware goal forecasting, ultimately resulting in smoother, collision-free navigable trajectories. Further, we also demonstrated the applicability of our model to current vision-language models for rich scene understanding and generating a detailed navigation instruction with appropriate reasoning. 

\printbibliography
\end{document}